\documentclass{article}
\usepackage{spconf,amsmath,graphicx}
\usepackage{svg}
\usepackage{amsmath, amssymb}
\usepackage{bm}
\usepackage{float}
\usepackage{graphicx}
\usepackage{caption}
\usepackage{subfig}
\usepackage{multirow}
\usepackage{algorithm2e}
\usepackage[utf8]{inputenc}
\usepackage{enumitem}
\usepackage{fancyvrb}
\usepackage{relsize}
\usepackage{tipa}
\usepackage{tikz}
\usepackage{pgfplots}
\usepackage{booktabs}
\usepackage{hyperref}
\usepackage{newunicodechar}
\usepackage[T1]{fontenc}
\newunicodechar{ˈ}{"}
\newunicodechar{ə}{@}
\newunicodechar{ɛ}{E}
\newunicodechar{ɪ}{I}
\newunicodechar{ʀ}{R}

\usetikzlibrary{positioning,arrows}

\newcommand{\norm}[1]{\left\lVert#1\right\rVert}

\title{Universal Phone Recognition with a Multilingual Allophone System}
\name{%
\begin{tabular}{@{}c@{}c@{}}
$^{\dagger}$Xinjian Li \qquad 
$^{\dagger}$Siddharth Dalmia \qquad 
$^{\dagger}$Juncheng Li\\ 
$^{\circ}$Matthew Lee \qquad 
$^{\triangle}$Patrick Littell \qquad 
$^{\square}$Jiali Yao \qquad
$^{\dagger}$Antonios Anastasopoulos \\
$^{\dagger}$David R. Mortensen \qquad 
$^{\dagger}$Graham Neubig \qquad 
$^{\dagger}$Alan W Black \qquad
$^{\dagger}$Florian Metze
\end{tabular}}

\address{$^{\dagger}$Carnegie Mellon University;
$^{\circ}$SIL International; \\
$^{\triangle}$National Research Council Canada; $^{\square}$ByteDance AI Lab \\
\\
\texttt{xinjianl@cs.cmu.edu}}

\begin{document}
\ninept

\maketitle

\begin{abstract}                                                          
Multilingual models can improve language processing, particularly for low resource situations, by sharing parameters across languages.
Multilingual acoustic models, however, generally ignore the difference between phonemes (sounds that can support lexical contrasts in a \emph{particular} language)
and their corresponding phones (the sounds that are actually spoken, which are language independent).
This can lead to performance degradation when combining a variety of training languages, as identically annotated phonemes can actually correspond to several different underlying phonetic realizations.
In this work, we propose a joint model of both language-independent phone and language-dependent phoneme distributions.
In multilingual ASR experiments over 11 languages, we find that this model improves testing performance by 2\% phoneme error rate absolute in low-resource conditions.
Additionally, because we are explicitly modeling language-independent phones, we can build a (nearly-)universal phone recognizer
that, when combined with the PHOIBLE~\cite{phoible} large, manually curated database of phone inventories, can be customized into 2,000 language dependent recognizers.
Experiments on two low-resourced indigenous languages, Inuktitut and Tusom, show that our recognizer achieves phone accuracy improvements of more than 17\%, moving a step closer to speech recognition for all languages in the world.\footnote{A web demo is available at \url{https://www.dictate.app}, the pretrained model will be released at \url{https://github.com/xinjli/allosaurus}}




\end{abstract}
\section{Introduction}
\label{sec:intro}



\begin{figure}[t]
  \centering
\begin{tikzpicture}[>=stealth, node distance=5mm and 1cm]
\node (peak_orth) {\itshape peak};
\node (speak_orth) [right = of peak_orth] {\itshape speak};
\node (ping_orth) [right = of speak_orth] {\itshape ping};
\node (bing_orth) [right = of ping_orth] {\itshape bing};
\node[xshift=1cm] (english) [above = 0.1\baselineskip of peak_orth] {\scshape English};
\node[xshift=1cm] (mandarin) [above = 0.1\baselineskip of ping_orth] {\scshape Mandarin Chinese};
\node (ping_gloss) [below = 0.1\baselineskip of ping_orth] {`level’};
\node (bing_gloss) [below = 0.1\baselineskip of bing_orth] {`ice’};
\node (peak_phoneme) [below = of peak_orth] {/\textipa{{\bfseries p}ik}/};
\node (speak_phoneme) [below = of speak_orth] {/\textipa{s{\bfseries p}ik}/};
\node (ping_phoneme) [below = of ping_orth] {/\textipa{{\bfseries p\super h}iN}/};
\node (bing_phoneme) [below = of bing_orth] {/\textipa{{\bfseries p}iN}/};
\node (peak_phone) [below = of peak_phoneme] {[\textipa{{\bfseries p\super h}ik}]};
\node (speak_phone) [below = of speak_phoneme] {[\textipa{s{\bfseries p}ik}]};
\node (ping_phone) [below = of ping_phoneme] {[\textipa{{\bfseries p\super h}iN}]};
\node (bing_phone) [below = of bing_phoneme] {[\textipa{{\bfseries p}iN}]};
\draw[<->] (peak_phoneme.south) to [bend right=10] (bing_phoneme.south);
\draw[<->] (speak_phoneme.south) to [bend right=10] (bing_phoneme.south);
\draw[<->] (peak_phone.south) to [bend right=10] (ping_phone.south);
\draw[<->] (speak_phone.south) to [bend right=10] (bing_phone.south);
\end{tikzpicture}
  \caption{Words, phonemes (slashes), and phones (square brackets).}
\label{example}
\end{figure}
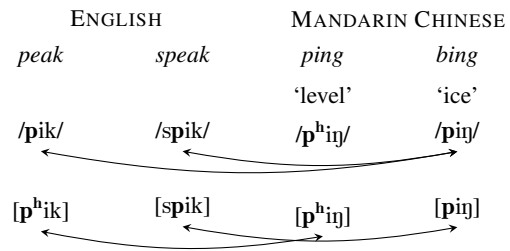

There is an increasing interest in building speech tools benefiting low-resource languages, specifically multilingual models that can improve low-resource recognition using rich resources available in other languages like English and Mandarin. One standard tool for recognition in low resource languages is multilingual acoustic modeling~\cite{dalmia2018sequence}. 
Acoustic models are generally trained on parallel data of speech waveforms and \emph{phoneme} transcriptions.
Importantly, phonemes are perceptual units of sound that closely correlate with, but do not exactly correspond to the actual sounds that are spoken, \emph{phones}.
An example of this is shown in Figure~\ref{example}, which demonstrates two English words that share the same phoneme /\textipa{p}/, but different in the actual phonetic realizations [\textipa{p}] and [\textipa{p\super h}].
\emph{Allophones}, the sets of phones that correspond to a particular phoneme, are language specific; distinctions that are important in some languages are not important in others.

Most multilingual acoustic models simply use existing phoneme transcriptions as-is, taking the union of the phoneme sets to be shared by all training languages \cite{lin2009study,cohen1997towards,schultz2001language,schultz1997fast,li2020zero}.
The assumption is reasonable under some circumstances as phoneme names are typically associated with their most common or least marked allophone. 
However, this is obviously an over-simplistic view: in Figure \ref{example}, for example, this would mean that all training in English would assign the phones [\textipa{p}] and [\textipa{p\super h}] to phoneme /\textipa{p}/. This is detrimental if we want to recognize Mandarin Chinese, for instance, where the two phones are corresponding to two distinctive phonemes /\textipa{p}/ and /\textipa{p\super h}/.

In this paper, we propose a novel method for multilingual recognition based on phonetic annotation to tackle this problem: \textit{Allosaurus} (\textbf{allo}phone \textbf{s}ystem of \textbf{au}tomatic \textbf{r}ecognition for \textbf{u}niversal \textbf{s}peech).
Our method incorporates knowledge of phonology into the multilingual model through an \textit{allophone layer}, which associates a universal narrow phone set with the phonemes that appear in the transcription of each language.
Our model first computes the phone distribution using a standard ASR encoder, then the allophone layer maps the phone distribution into the phoneme distribution for each language.
This model can be trained end-to-end using only standard phonemic transcriptions and an allophone list created by phoneticians.
The allophone layer is first initialized with the allophone list, then is further optimized during the training process.
We demonstrate that accounting for the phoneme-phone mismatch in this way improves the accuracy of multilingual acoustic modeling by 2.0\% phoneme error rate in low-resource conditions.


Furthermore, the architecture simultaneously makes it possible to perform \emph{universal phone recognition}.
Previous approaches cannot perform phone recognition in a universal fashion as they depend on language-specific phonemes, as illustrated with the previous example of English not distinguishing /\textipa{p}/ and /\textipa{p \super h}/ as required in Mandarin.
In contrast, because our approach allows recognition of phones directly, it already has learned to make these fine-grained distinctions.
Taking advantage of this fact, we incorporate a large phone inventory database collected by linguists, PHOIBLE \cite{phoible}, and demonstrate that our phone recognizer can be customized to recognize over 2000 languages without any training data in the languages themselves.
By evaluating the recognizer with completely unseen testing languages, we found that our recognizer achieves 17\% better performance absolute compared with the traditional approach. 




\section{Related Work}
\label{sec:related}

While some recent work in multilingual ASR focuses on end-to-end models to directly predict graphemes~\cite{watanabe2017language, toshniwal2018multilingual}, most systems still depend on phonetically inspired acoustic models. Multilingual acoustic models fall into two groups.
The first group, \emph{shared phoneme} models, creates a shared phoneme inventory of all phonemes from all training languages~\cite{lin2009study,cohen1997towards,schultz2001language,schultz1997fast,li2020zero, thompson2019transferable}.
The second group, \emph{private phoneme} models, treats phonemes from each language as completely different classes performs phoneme classification separately for each language~\cite{huang2013cross,dalmia2018sequence,li2019multilingual}.
However, these two groups have their own respective drawbacks: the first group fails to consider the disconnect between the phonemes across languages while the second group completely ignores cross-lingual phonetic associations and is not applicable to recognition of new languages. In contrast, our approach solves both of these problems by taking into account allophones with phone-phoneme mappings.

There have been some attempts to apply phone recognizers to low resource languages. For example, English recognizers have been applied to align transcription corpora of an endangered language \cite{dicanio2013using}, facilitate language documentation \cite{michaud2018integrating}, identify languages with language models \cite{matejka2005phonotactic}, and perform linguistic annotation \cite{neubig2018towards}. However, these approaches depend heavily on training data in the language of interest and their specific phonemic transcriptions. Our approach, on the other hand, abstracts away the dependency to phonemes by applying the allophone transformations.

\section{Approach}

\begin{figure}[t]
  \centering
  \includegraphics[width=\columnwidth]{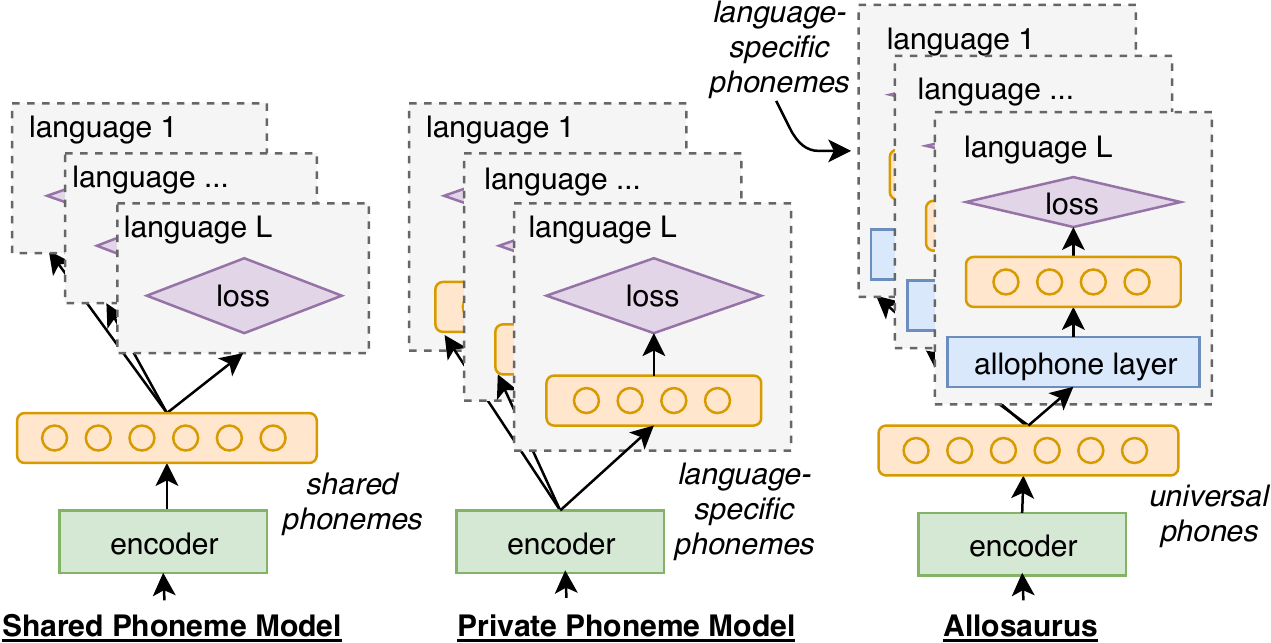}
  \caption{Traditional approaches predict phonemes directly, either for all languages (left) or separately for each language (middle). On the contrary, our approach (right) predicts over a shared phone inventory, then maps into language-specific phonemes with an allophone layer.}
  \vspace{-3mm}
\label{arch}
\end{figure}

\subsection{Phone-Phoneme Annotation}
Suppose there are $|L|$ training languages, and each language $L_i$ has its own phoneme inventory $Q_{i}$ which can be easily obtained by enumerating the phonemes appearing in its annotated training data.
Most traditional multilingual approaches handle inventories at the phoneme level, and create a \textit{shared phoneme inventory} $Q_{\text{sha}}$ by taking union of the phoneme sets:

\begin{equation}
    \small
    Q_{\text{sha}} = \bigcup_{1 \leq i \leq |L| } Q_i.
\end{equation}

In contrast, our method distinguishes phonemes from their phone realizations.
We have linguists annotate each phoneme $q \in Q_{i}$ with its corresponding allophone set $P^i_q$, where each phone $p \in P^i_q$ is a realization of $q$ in language $L_i$.
Merging these sets for all languages, we obtain the \textit{universal phone inventory} $P_{\text{uni}}$.

\begin{equation}
    \small
    P_{\text{uni}} = \bigcup_{1 \leq i \leq |L| } \bigcup_{q \in Q_{i}} P^i_q
\end{equation}

Additionally, we obtain a \textit{signature matrix} $S^i = \{0, 1\}^{|Q_i| \times |P_{\text{uni}}|} $ describing the association of phone and phonemes in each language $L_i$: Suppose the phoneme $q \in Q_i$ has the row index $j$ where $1 \leq j \leq |Q_i|$ , phone $p \in P_{\text{uni}}$ has the column index $k$ where $1 \leq k \leq |P_{\text{uni}}|$, if the $p$ is a realization of $q$, then $(j,k)$ cell of the $S^i$ has a value of 1, otherwise it is assigned 0.


\subsection{Allophone Layer}

As mentioned in Section~\ref{sec:related}, traditional multilingual models can be divided into two groups.
The first group, \emph{shared phoneme models} (Figure \ref{arch} left), predicts phoneme distributions over the shared phoneme inventory $Q_{\text{sha}}$. The second group, \emph{private phoneme models} (Figure \ref{arch} middle), on the other hand, shares a common encoder but computes distribution over private phoneme inventory $Q_i$ for each language $L_i$. These approaches handle phonemes directly with no concept of underlying phones.

\begin{table*}[t!]
\begin{center}
    \caption{Results of three models' phoneme error rate performance on 11 languages. The top-half shows the results trained with all training datasets. The bottom-half shows the low-resource results in which only 1k utterances are used for training from each dataset.} \label{tab:training_results} 
    \begin{tabular}{l l | c c c c c c c c c c c | c} 
    \toprule
    & & Amh & Eng & Ger & Ita & Jap & Man & Rus & Spa & Tag & Tur & Vie & Average \\
    \midrule
    \multirow{3}{*}{\rotatebox{90}{\textbf{Full}}} & \textbf{Shared Phoneme PER}  & 78.4 & 71.7 & 71.6 & 62.9 & 65.9 & 76.5 & 76.9 & 62.6 & 74.1 & 76.6 & 82.7 & 73.8 \\
    & \textbf{Private Phoneme PER} & 37.1 & 22.4 & 17.6 & 26.2 & 17.6 & 17.9 & 21.3 & 18.5 & 47.6 & 35.8 & 56.5 & 25.6 \\
    & \textbf{Allosaurus PER} & 36.0 & 20.5 & 18.8 & 23.7 & 23.8 & 17.0 & 26.3 & 19.4 & 57.4 & 35.3 & 57.3 & 25.0 \\
    \midrule
    \multirow{3}{*}{\rotatebox{90}{\textbf{Low}}} & \textbf{Shared Phoneme PER} & 80.4 & 73.3 & 74.3 & 72.2  & 77.1 & 83.0 & 83.2 & 72.8 & 84.8 & 84.4 & 84.5 & 78.4 \\
    & \textbf{Private Phoneme PER} & 55.4 & 50.6 & 41.9 & 31.6 & 36.8 & 37.0 & 47.9 & 36.7 & 62.3 & 54.5 & 73.6 & 43.8 \\
    & \textbf{Allosaurus PER} & 54.8 & 47.0 & 41.5 & 37.4 & 40.5 & 33.4 & 45.0 & 35.9 & 70.1 & 53.6 & 72.5 & 41.8 \\
    \bottomrule
    \end{tabular}
    \vspace{-2mm}
\end{center}
\end{table*}

\begin{table}[t!]
\begin{center}
    \caption{Training corpora and size in utterances for each language. Models are trained and tested with 12 rich resource languages (top) and 2 low resource unseen languages (bottom).} \label{tab:corpus}
    \resizebox{\columnwidth}{!}{
    \begin{tabular}{l l r} 
    \toprule
    {\bf Language} & {\bf Corpora} & {\bf Utt.} \\
    \midrule
    English & voxforge, Tedlium \cite{rousseau2012ted}, Switchboard \cite{godfrey1992switchboard} & 1148k \\
    Japanese & Japanese CSJ \cite{maekawa2003corpus} &  440k \\
    Mandarin & Hkust \cite{liu2006hkust}, openSLR \cite{aishell_2017,THCHS30_2015} & 377k \\
    Tagalog & IARPA-babel106b-v0.2g & 93k \\
    Turkish & IARPA-babel105b-v0.4 & 82k \\
    Vietnamese & IARPA-babel107b-v0.7 & 79k \\
    German & voxforge & 40k \\
    Spanish & LDC2002S25 & 32k \\
    Amharic & openSLR25 \cite{Abate2005} & 10k \\
    Italian & voxforge & 10k \\
    Russian & voxforge & 8k \\
    \midrule
    Inukitut & private & 1k \\
    Tusom & private & 1k \\
    \bottomrule
    \end{tabular}
    }
    \vspace{-2mm}
\end{center}
\end{table}

In contrast our proposed approach, \emph{Allosaurus},  (Figure \ref{arch} right), comprises a language independent encoder and phone predictor, and a language dependent allophone layer and a loss function associated with each language. The encoder first produces the distribution $h \in \mathbb{R}^{|P_{\text{uni}}|}$ over the universal phone inventory $P_{\text{uni}}$, then the allophone layer transforms $h$ into phoneme distribution $g^i \in \mathbb{R}^{|Q_{i}|}$ of each language. 
The allophone layer uses a trainable allophone matrix $W^i \in \mathbb{R}^{|Q_i| \times |P_{\text{uni}}|}$ to describe allophones in the similar way as $S^i$. The allophone matrix $W^i$ is first initialized with $S^i$, and is allowed to be optimized during the training process, but we add an L2 penalty to penalize divergence from the original signature matrix $S^i$. The allophone layer computes its logit distribution $g^i$ by finding the most likely allophone realization in $P_{\text{uni}}$ with maxpooling. 






\begin{equation}
g^i_{j}=\max(\{ w^i_{j,k} \cdot h_{k} ; 1 \leq k \leq |P_{\text{uni}}| \}),
\end{equation}
where $g^i_{j} \in \mathbb{R}$ is the logit of $j$-th phoneme in $g^i$ for language $L_i$, $w^i_{j,k} \in \mathbb{R}$ is the $(j,k)$ cell of the allophone matrix $W^i$, $h_{k} \in \mathbb{R}$ is the logit of $k$-th phone in $h$. Intuitively, if the $j$-th phoneme has the $k$-th phone as an allophone, $w^i_{j,k}$ would be near 1, otherwise $w^i_{j,k}$ would be near 0. Therefore, the phoneme logit of $g^i_j$ is decided by the largest allophone logit $h_k$. The phoneme distribution $g^i$ is further fed into the loss function. This method for phoneme prediction can be used with any underlying multilingual ASR system. Here we specifically optimize the parameters by minimizing CTC loss \cite{graves2006connectionist} for all training languages, with the addition of regularization of the allophone layer controlled by hyperparameter $\alpha$. 


\begin{equation}
\mathcal{L}=\sum_{1 \leq i \leq |L|} (\mathcal{L}^i_{ctc} + \alpha \norm{ W^i-S^i }_2^2 ).
\vspace{-2mm}
\end{equation}

\subsection{Universal Phone Recognition}

Not only does the allophone layer abstract away from the language-specific phonemes, which contributes to the improvement in the multilingual acoustic modeling, the model also gives us the capability to predict universal phones themselves.
This has rarely been attempted in previous work.
By applying the greedy decoding strategy over the phone distribution $h$, we can obtain a phone sequence in which all phones $P_{\text{uni}}$ in the training languages are candidates. When combined with a large training languages sets, our universal inventory is expected to cover most common narrow phones appearing in many languages in the world, which we show in the experiment section. 

Furthermore, this recognition protocol can take into account phone inventories that have already been created for many languages in the world by linguists.
For example, PHOIBLE \cite{phoible} is a database of phone inventories for more than 2000 languages and dialects, allowing our model to be applied to these languages with some degree of accuracy.
If the phone inventory for language $L_i$ is $P_i$, we can restrict the decoder to only produce phones in $P_i \cap P_{\text{uni}}$ by filtering out other phones.
When the universal inventory $P_{\text{uni}}$ covers most frequent phones in the world, we could expect that $P_i \approx P_i \cap P_{\text{uni}}$.





\section{Experiments}

\subsection{Settings}
As we are interested in creating a large universal phone inventory, we select a phonetically diverse set of 11 training languages as summarized on the top of Table~\ref{tab:corpus}. We include corpora from a variety of speech domains to make our model robust (e.g., read speech, sponatenuous speech). 5\% of the dataset is used as the test set, and the remaining data are used as the training set and the validation set. We also consider a low resource condition, where 1,000 random utterances are used from each corpus to train the model. As baselines, we compare with the previously-described \textit{shared phoneme} and \textit{private phoneme} models.
All methods use the same encoder and features.
Features are high-resolution 40 dimensional MFCCs extracted with Kaldi~\cite{povey2011kaldi}. The encoder is a 6-layer stacked bidirectional LSTM with hidden size of 1024 in each layer. The regularization hyperparameter $\alpha$ is set to 10.
Phonemes for training languages are assigned using the grapheme-to-phoneme tool Epitran \cite{mortensen2018epitran}.
For each phoneme in each language, phoneticians (mostly authors of this paper) create the allophone mappings.\footnote{This work has been accepted to LREC 2020 and its database is available at \url{https://github.com/dmort27/allovera}}

We evaluate using phoneme error rate for the training languages.
Furthermore, we select two languages not included in the training data: Inukitut and Tusom.
These languages are indigenous languages with few training resources, representing a realistic scenario where our model is applied to entirely new languages, as may be done when ASR is used for documentation of endangered languages.
The datasets of these two languages are transcribed with phones, and accordingly we use phone error rate rather than the phoneme error rate.
While Allosaurus is able to predict phones in a natural way by decoding $h$, the two baselines could not predict phones directly. In this unseen language experiment, we assume phonemes predicted by the baselines correspond to phones of the same name.

\begin{table}[t]
\begin{center}
    \caption{Statistics of the phone coverage mean (standard deviation) of areas. Phone coverage of language $L_i$ is defined as $\frac{| P_{\text{uni}} \cap P_{i}|}{|P_{i}|}$} \label{tab:family}
    \begin{tabular}{l r r r } 
    \toprule
    {\bf Area} & {\bf \# Language} & {\bf Shared} & {\bf Allosaurus} \\
    \midrule
     Africa & 875 & 53\% (13\%)&  84\% (11\%) \\
     America & 659 & 52\% (14\%) & 81\% (13\%) \\
    Asia & 377 & 46\% (15\%) & 79\% (13\%) \\
    Pacific & 152 & 59\% (15\%) & 87\% (12\%) \\
    Europe & 92 & 35\% (9.5\%) & 69\% (13\%) \\
    \midrule
    All & 2155 & 52\% (15\%) & 82\% (13\%) \\

    \bottomrule
    \end{tabular}
    \vspace{-2mm}
\end{center}
\end{table}

\begin{table}[t]
\begin{center}
    \caption{Comparisons of phone error rates in two unseen languages} \label{tab:unseen_results}
    \begin{tabular}{l r r} 
    \toprule
    & Inuktitut & Tusom \\
    \midrule
    \textbf{Shared Phoneme PER} & 94.1 & 93.5 \\
    \textbf{Best Private Phoneme PER} & 86.2 & 85.8 \\
    \midrule
    \textbf{Allosaurus PER} & 84.1 & 77.3  \\
    \textbf{Allosaurus+PHOIBLE PER} & 73.1 & 64.2 \\
    \bottomrule
    \end{tabular}
    \vspace{-2mm}
\end{center}
\end{table}

\subsection{Main Results}


Table~\ref{tab:training_results} demonstrates the performance of the baseline models and Allosaurus evaluated on 11 languages. The top half of the table summarizes the performance when trained with the full training set. The results suggests both the private phoneme model and the Allosaurus model outperforms the shared phoneme model significantly. The results of the shared phoneme model can be explained by the disagreement of phoneme assignments across languages. In contrast, the private phoneme model handles this issue by using language specific phoneme layers. Our model also circumvents this issue by introducing the language-specific allophone layers. The bottom half of the Table~\ref{tab:training_results} highlights the results when the training set of each language is limited as mentioned above. Unsurprisingly, limiting the amount of training data hurts accuracy across the board. While the private phoneme model and our model achieve similar results when using the full training set, our model outperforms the private phoneme model by 2.0\% when training data is limited. This suggests that our model is better at sharing parameters across languages by using prior phonetic knowledge in this case, likely due to the fact that the private phoneme model needs to learn each phoneme predictor from scratch, while our model already has phone-phoneme mapping knowledge seeded by linguistically motivated annotations.

\begin{table}[t]
\begin{center}
    \caption{An English example from switchboard in which Allosaurus could distinguish [\textipa{p\super h}] and [\textipa{p}] for phoneme /\textipa{p}/} \label{tab:english_res}
    \begin{tabular}{l l} 
    \toprule
    Model & Phones  \\
    \midrule
    \textbf{Utterance} & the quebec \textbf{people} that that \textbf{speak} french \\
    \textbf{Annotation} & /\dh \, \textipa{@} \textipa{k} \textipa{w} \textipa{@} b \textipa{E} k \textbf{\textipa{p}} \textipa{i} \textipa{p} \textipa{@} \textipa{l} .. \textipa{s} \textbf{\textipa{p}} \textipa{i} \textipa{k} \textipa{f} \textturnr\, \textipa{E} \textipa{n} \textteshlig/ \\
    \textbf{Allosaurus} & [\dh \, \textipa{@} \textipa{x} \textipa{o} \textipa{b} \textipa{@} \textipa{k} \textturna \,
 \textbf{\textipa{p\super h}} \textipa{i} \texttheta\, \textipa{o:} \textipa{l} .. s \textbf{p} \textturnr\, \textipa{I} \textipa{k} \textipa{f} \textturnr\, \textipa{E} \textipa{n} \textipa{d}] \\
    \bottomrule
    \end{tabular}
    \vspace{-2mm}
\end{center}
\end{table}



\subsection{Universal Phone Recognition Results}

In addition to the improvements over low resource settings, our model enables us to predict (nearly-)universal phone distributions.
By merging phone inventories from all of our languages, we obtain a shared inventory of 187 phones. First, we assess how close this inventory gets to covering the languages registered in PHOIBLE.
The Allosaurus column in Table~\ref{tab:family} summarizes the phone coverage of our model, split into different geographic areas.
The phone coverage in each cell represents the mean and standard deviation for each category. As the table suggests, our model has a promising phone coverage over all areas consistently. On average, it has 82\% mean phone coverage and 12.8\% standard deviation over all PHOIBLE languages.
Furthermore, by comparing our model with the baseline model in which we merged all the phoneme inventories from the corpus as-is, we significantly improve the phone coverage by 30\%. Additionally, the standard deviation shows that our model covers phones more consistently than the baseline model.

Next, we actually evaluate the model with respect to its ability to recognize phones. Table \ref{tab:english_res} shows a decoded English example. The utterance contains three English phonemes /p/ in word \textit{people} and \textit{speak}. The underlying allophones, however, are [\textipa{p\super h}] and [\textipa{p}] as mentioned in Section 1. While the original English training set annotates those two words with the same phoneme /p/, Allosaurus is able to predict different allophones by leveraging knowledge from other languages (e.g: Mandarin). We also note that Allosaurus is still not perfect: it fails to recognize the second /p/ in ``people.''

Additionally we also investigate unseen languages on the Inuktitut and Tusom datasets. The results are summarized in the Table~\ref{tab:unseen_results}. As the result show, the shared phoneme model can hardly recognize any phonemes in these two languages, with more than 90.0\% phone error rate on both datasets.
Next, we try all 11 private phoneme models from the training datasets and use the one with the lowest phoneme error rate. 
Unsurprisingly, this also can not achieve satisfying results on both datasets, as none of our 11 languages is similar to Inuktitut and Tusom; they both have over 85.0\% phone error rate.
On the other hand, the proposed Allosaurus model achieves 84.1\% phone error rate on Inuktitut and 77.3\% phone error rate on Tusom, a significant drop.
When combined with the PHOIBLE inventory, the error rates are further improved to 73.1\% and 64.2\% respectively, which shows 17\% improvements on average over the shared phoneme baseline. Table \ref{tab:inuktitut_res} shows one qualitative example from Inuktitut data. It suggests that simply applying Allosaurus could capture some aspects of the target phonemes, but it still made many errors especially substitution errors between [e] and [i]. The reason is Allosaurus has a much broader phone search space (187 phones), it might be difficult to distinguish similar phones (e.g: both [e] and [i] are front vowels, but [e] is a close vowel and [i] is a close-mid vowel). We find those substitution errors account for the majority of errors in the test sets. Those confusing phones, however, might be solved when combined with an appropriate inventory such as PHOIBLE. The last row suggests that Allosaurus could fix those substitution errors as [e] does not exist in Inuktitut's inventory. 


\begin{table}[t]
\begin{center}
    \caption{A qualitative example from Inuktitut dataset} \label{tab:inuktitut_res}
    \begin{tabular}{l l} 
    \toprule
    Model & Phones  \\
    \midrule
    \textbf{Ground Truth} & [\textipa{i} \textipa{l} \textipa{i} \textipa{t} \textipa{s} \textipa{i} \textipa{l:} \textipa{i}] \\
    \textbf{Allosaurus} & [\textipa{e} \textipa{l} \textipa{e} \textipa{p} \textscr\, \textipa{I} \textipa{l:} \textipa{e}] \\
    \textbf{Allosaurus+PHOIBLE} & [\textipa{i} \textipa{l} \textipa{i} \textipa{t} \textipa{i} \textipa{l:} \textipa{i:}] \\
    \bottomrule
    \end{tabular}
    \vspace{-2mm}
\end{center}
\end{table}


\section{Conclusion}
In this work, we propose \textit{Allosaurus}, which considers the relationship between phones and phonemes in multilingual acoustic modeling. 
It improves significantly the phone recognition accuracy over unseen languages by 17\%.

\section{Acknowledgment}
This work is supported by NSF awards ACI-1548562 and 1761548. We would like to thank Alexis Michaud, Steven Abney, Hilaria Cruz and other participants of the LTLDR workshop for their feedback.

\bibliographystyle{IEEEbib}
\bibliography{refs}

\begin{thebibliography}{10}

\bibitem{phoible}
Steven Moran and Daniel McCloy, Eds.,
\newblock {\em PHOIBLE 2.0},
\newblock Max Planck Institute for the Science of Human History, Jena, 2019.

\bibitem{dalmia2018sequence}
Siddharth Dalmia, Ramon Sanabria, Florian Metze, and Alan~W Black,
\newblock ``Sequence-based multi-lingual low resource speech recognition,''
\newblock in {\em 2018 IEEE International Conference on Acoustics, Speech and
  Signal Processing (ICASSP)}. IEEE, 2018, pp. 4909--4913.

\bibitem{lin2009study}
Hui Lin, Li~Deng, Dong Yu, Yi-fan Gong, Alex Acero, and Chin-Hui Lee,
\newblock ``A study on multilingual acoustic modeling for large vocabulary
  asr,''
\newblock in {\em 2009 IEEE International Conference on Acoustics, Speech and
  Signal Processing}. IEEE, 2009, pp. 4333--4336.

\bibitem{cohen1997towards}
Paul~S. Cohen, Satyanarayana Dharanipragada, Jerneja~Zganec Gros,
  Michael~Daniel Monkowski, Chalapathy Neti, Salim Roukos, and Todd Ward,
\newblock ``Towards a universal speech recognizer for multiple languages,''
\newblock in {\em 1997 IEEE Workshop on Automatic Speech Recognition and
  Understanding Proceedings}. IEEE, 1997, pp. 591--598.

\bibitem{schultz2001language}
Tanja Schultz and Alex Waibel,
\newblock ``Language-independent and language-adaptive acoustic modeling for
  speech recognition,''
\newblock {\em Speech Communication}, vol. 35, no. 1-2, pp. 31--51, 2001.

\bibitem{schultz1997fast}
Tanja Schultz and Alex Waibel,
\newblock ``Fast bootstrapping of lvcsr systems with multilingual phoneme
  sets,''
\newblock in {\em Fifth European Conference on Speech Communication and
  Technology}, 1997.

\bibitem{li2020zero}
Xinjian Li, Siddharth Dalmia, David~R Mortensen, Juncheng Li, Alan~W Black, and
  Florian Metze,
\newblock ``Towards zero-shot learning for automatic phonemic transcription,''
\newblock in {\em Thirty-Fourth AAAI Conference on Artificial Intelligence},
  2020.

\bibitem{watanabe2017language}
Shinji Watanabe, Takaaki Hori, and John~R Hershey,
\newblock ``Language independent end-to-end architecture for joint language
  identification and speech recognition,''
\newblock in {\em 2017 IEEE Automatic Speech Recognition and Understanding
  Workshop (ASRU)}. IEEE, 2017, pp. 265--271.

\bibitem{toshniwal2018multilingual}
Shubham Toshniwal, Tara~N Sainath, Ron~J Weiss, Bo~Li, Pedro Moreno, Eugene
  Weinstein, and Kanishka Rao,
\newblock ``Multilingual speech recognition with a single end-to-end model,''
\newblock in {\em 2018 IEEE International Conference on Acoustics, Speech and
  Signal Processing (ICASSP)}. IEEE, 2018, pp. 4904--4908.

\bibitem{thompson2019transferable}
Jessica~AF Thompson, Marc Sch{\"o}nwiesner, Yoshua Bengio, and Daniel Willett,
\newblock ``How transferable are features in convolutional neural network
  acoustic models across languages?,''
\newblock in {\em ICASSP 2019-2019 IEEE International Conference on Acoustics,
  Speech and Signal Processing (ICASSP)}. IEEE, 2019, pp. 2827--2831.

\bibitem{huang2013cross}
Jui-Ting Huang, Jinyu Li, Dong Yu, Li~Deng, and Yifan Gong,
\newblock ``Cross-language knowledge transfer using multilingual deep neural
  network with shared hidden layers,''
\newblock in {\em 2013 IEEE International Conference on Acoustics, Speech and
  Signal Processing}. IEEE, 2013, pp. 7304--7308.

\bibitem{li2019multilingual}
Xinjian Li, Siddharth Dalmia, Alan~W Black, and Florian Metze,
\newblock ``Multilingual speech recognition with corpus relatedness sampling,''
\newblock {\em Proc. Interspeech 2019}, pp. 2120--2124, 2019.

\bibitem{dicanio2013using}
Christian DiCanio, Hosung Nam, Douglas~H Whalen, H~Timothy~Bunnell, Jonathan~D
  Amith, and Rey~Castillo Garc{\'\i}a,
\newblock ``Using automatic alignment to analyze endangered language data:
  Testing the viability of untrained alignment,''
\newblock {\em The Journal of the Acoustical Society of America}, vol. 134, no.
  3, pp. 2235--2246, 2013.

\bibitem{michaud2018integrating}
Alexis Michaud, Oliver Adams, Trevor~Anthony Cohn, Graham Neubig, and
  S{\'e}verine Guillaume,
\newblock ``Integrating automatic transcription into the language documentation
  workflow: Experiments with na data and the persephone toolkit,''
\newblock 2018.

\bibitem{matejka2005phonotactic}
Pavel Matejka, Petr Schwarz, Jan Cernock{\`y}, and Pavel Chytil,
\newblock ``Phonotactic language identification using high quality phoneme
  recognition,''
\newblock in {\em Ninth European Conference on Speech Communication and
  Technology}, 2005.

\bibitem{neubig2018towards}
Graham Neubig, Patrick Littell, Chian-Yu Chen, Jean Lee, Zirui Li, Yu-Hsiang
  Lin, and Yuyan Zhang,
\newblock ``Towards a general-purpose linguistic annotation backend,''
\newblock {\em arXiv preprint arXiv:1812.05272}, 2018.

\bibitem{rousseau2012ted}
Anthony Rousseau, Paul Del{\'e}glise, and Yannick Esteve,
\newblock ``{TED-LIUM}: an automatic speech recognition dedicated corpus.,''
\newblock in {\em LREC}, 2012, pp. 125--129.

\bibitem{godfrey1992switchboard}
John~J Godfrey, Edward~C Holliman, and Jane McDaniel,
\newblock ``{SWITCHBOARD}: Telephone speech corpus for research and
  development,''
\newblock in {\em Acoustics, Speech, and Signal Processing, 1992. ICASSP-92.,
  1992 IEEE International Conference on}. IEEE, 1992, vol.~1, pp. 517--520.

\bibitem{maekawa2003corpus}
Kikuo Maekawa,
\newblock ``Corpus of spontaneous japanese: Its design and evaluation,''
\newblock in {\em ISCA \& IEEE Workshop on Spontaneous Speech Processing and
  Recognition}, 2003.

\bibitem{liu2006hkust}
Yi~Liu, Pascale Fung, Yongsheng Yang, Christopher Cieri, Shudong Huang, and
  David Graff,
\newblock ``Hkust/mts: A very large scale mandarin telephone speech corpus,''
\newblock in {\em Chinese Spoken Language Processing}, pp. 724--735. Springer,
  2006.

\bibitem{aishell_2017}
Xingyu Na Bengu Wu Hao~Zheng Hui~Bu, Jiayu~Du,
\newblock ``Aishell-1: An open-source mandarin speech corpus and a speech
  recognition baseline,''
\newblock in {\em Oriental COCOSDA 2017}, 2017, p. Submitted.

\bibitem{THCHS30_2015}
Zhiyong~Zhang Dong~Wang, Xuewei~Zhang,
\newblock ``Thchs-30 : A free chinese speech corpus,'' 2015.

\bibitem{Abate2005}
Solomon~Teferra Abate, Wolfgang Menzel, and Bairu Tafila,
\newblock ``An amharic speech corpus for large vocabulary continuous speech
  recognition,''
\newblock in {\em INTERSPEECH-2005}, 2005.

\bibitem{graves2006connectionist}
Alex Graves, Santiago Fern{\'a}ndez, Faustino Gomez, and J{\"u}rgen
  Schmidhuber,
\newblock ``Connectionist temporal classification: labelling unsegmented
  sequence data with recurrent neural networks,''
\newblock in {\em Proceedings of the 23rd international conference on Machine
  learning}. ACM, 2006, pp. 369--376.

\bibitem{povey2011kaldi}
Daniel Povey, Arnab Ghoshal, Gilles Boulianne, Lukas Burget, Ondrej Glembek,
  Nagendra Goel, Mirko Hannemann, Petr Motlicek, Yanmin Qian, Petr Schwarz,
  et~al.,
\newblock ``The kaldi speech recognition toolkit,''
\newblock in {\em IEEE 2011 workshop on automatic speech recognition and
  understanding}. IEEE Signal Processing Society, 2011, number CONF.

\bibitem{mortensen2018epitran}
David~R Mortensen, Siddharth Dalmia, and Patrick Littell,
\newblock ``Epitran: Precision {G2P} for many languages.,''
\newblock in {\em LREC}, 2018.

\end{thebibliography}

\end{document}